%% file: main.tex
\documentclass[10pt,twocolumn,letterpaper]{article}

\usepackage[pagenumbers]{cvpr} % To force page numbers, e.g. for an arXiv version

\input{preamble}

\definecolor{cvprblue}{rgb}{0.21,0.49,0.74}
\usepackage[pagebackref,breaklinks,colorlinks,citecolor=cvprblue]{hyperref}

\def\paperID{6831} % *** Enter the Paper ID here
\def\confName{CVPR}
\def\confYear{2024}

\title{LGFCTR: Local and Global Feature Convolutional Transformer for Image Matching}

\author{Wenhao Zhong\\
Beihang University\\
Beijing, China\\
{\tt\small zhongwenhao2022@buaa.edu.cn}
\and
Jie Jiang\\
Beihang University\\
Beijing, China\\
{\tt\small jiangjie@buaa.edu.cn}
}

\begin{document}
\maketitle
\input{sec/0_abstract}    
\input{sec/1_intro}
\input{sec/2_related}
\input{sec/3_method}
\input{sec/4_experiments}
\input{sec/5_conclusion}
\input{sec/X_suppl}
\clearpage
\clearpage
{
    \small
    \bibliographystyle{ieeenat_fullname}
    \bibliography{main}
}

\end{document}

%% file: preamble.tex
\usepackage[dvipsnames]{xcolor}

\usepackage{indentfirst}

\usepackage{multirow}
\usepackage[normalem]{ulem}
\useunder{\uline}{\ul}{}
\usepackage{makecell}

%% file: sec/0_abstract.tex
\begin{abstract}
Image matching that finding robust and accurate correspondences across images is a challenging task under extreme conditions. Capturing local and global features simultaneously is an important way to mitigate such an issue but recent transformer-based decoders were still stuck in the issues that CNN-based encoders only extract local features and the transformers lack locality. Inspired by the locality and implicit positional encoding of convolutions, a novel convolutional transformer is proposed to capture both local contexts and global structures more sufficiently for detector-free matching. Firstly, a universal FPN-like framework captures global structures in self-encoder as well as cross-decoder by transformers and compensates local contexts as well as implicit positional encoding by convolutions. Secondly, a novel convolutional transformer module explores multi-scale long range dependencies by a novel multi-scale attention and further aggregates local information inside dependencies for enhancing locality. Finally, a novel regression-based sub-pixel refinement module exploits the whole fine-grained window features for fine-level positional deviation regression. The proposed method achieves superior performances on a wide range of benchmarks. The code will be available on https://github.com/zwh0527/LGFCTR.
\end{abstract}

%% file: sec/1_intro.tex
\section{Introduction}
\label{sec:intro}

\input{figure/CNN_TR_CTR}

Image matching that finding correspondences across images sharing co-visible area lays the foundation for various geometric computer vision downstream tasks, including structure from motion (SFM)~\cite{sfm1, sfm2}, visual localization~\cite{loc1, HLoc, loc3}, simultaneous localization and mapping (SLAM)~\cite{slam1, slam2}, etc. Traditional methods like SIFT~\cite{orb, sift, surf, rootsift} extract keypoints and descriptors based on handcrafted features and generate matches by measuring the similarities among descriptors. Detector-based methods~\cite{SiLK, SCFeat, DISK, CAPS, GIFT, ASLFeat, R2D2, D2-Net, SP, PoSFeat, RF-Net, LF-Net, HN++, L2-Net, KeyNet, HAN, Contextdesc} still follow this pipeline but use learned features instead. They improve the performances but have poor repeatability of detector and reliability of description under extreme conditions such as large viewpoint and illumination changes, texture loss and repetitive patterns.

Detector-free methods~\cite{GOCor, PDC-Net, QuadTree, DKM, IFCAT, GLU-Net, DGC-Net, MatchFormer, AdaMatcher, AspanFormer, LoFTR, COTR, DualRC-Net, SparseNC-Net, NC-Net} as another line of works eschewed explicit keypoints extraction and generate correspondences directly from raw images in an end-to-end manner, which could take full advantages of rich contexts to achieve more accurate and robust matching than detector-based methods. NC-Net and its variants~\cite{NC-Net, SparseNC-Net, DualRC-Net} rely on iterative convolutions upon cost volumes to filter match information but lack long range dependencies. LoFTR and its variants~\cite{LoFTR, AdaMatcher, AspanFormer, COTR} mitigate such an issue by adopting transformer-based decoders but still only extract local features by CNN-based encoders as shown in \cref{fig:CNN_TR_CTR}(a). However, since human vision focuses on not only local visual appearances but also global reliances on features at other positions, the encoder should capture local and global features simultaneously. MatchFormer~\cite{MatchFormer} directly adopts a transformer-based encoder to extract global features. However, on one hand, explicit positional encoding leads transformers to fit into training resolutions, which hurts the generalization of transformers on varying testing resolutions. On the other hand, transformers capture few local features due to weak locality caused by point-based linear projections as shown in \cref{fig:CNN_TR_CTR}(b). In contrast, as shown in \cref{fig:CNN_TR_CTR}(a), convolution templates impose strong local spatial structure reliances on feature extraction, leading to strong locality and implicit spatial relative positional encoding.

Inspired by these two insights, a novel convolutional transformer without explicit positional encoding is proposed as a detector-free matcher and termed as Local and Global Feature Convolutional Transformer (LGFCTR) since it completely exploits transformers and convolutions to capture both local contexts and global structures more sufficiently. Specifically, the fusions between convolutions and transformers lay on there aspects: 1) A stem CNN for extracting shallow local textures is followed by a transformer-based self-encoder for deep feature extraction; 2) Both self-encoder and cross-decoder adopt transformers to capture long range dependencies for feature extraction at the same resolution and adopt convolutions to enhance locality for feature transitions across different resolutions; 3) As the basic block of encoder and decoder, the novel convolutional transformer module (CTR) compensates transformers for the locality of convolutions. More specifically, as shown in \cref{fig:CNN_TR_CTR}(c), on one hand, a novel multi-scale attention (MSA) firstly adopts area-based convolutions of multi kernel sizes instead of point-based linear projections to aggregate local contexts in different scales as a preprocessing, and then explores multi-scale long range dependencies between features of multi receptive fields during attention operations. On the other hand, a local pooling unit (LPU) as a postprocessing focuses on exploring local information inside long range dependencies in both spatial and channel dimensions as well as enhancing locality. 

In addition, fine-level matching refinement is always adopted to achieve sub-pixel matching accuracy. However, many detector-free methods~\cite{LoFTR, AspanFormer, AdaMatcher} simply adjust matching positions based on the expectations of relevant scores between the central feature of source window and the whole target window, which can’t take full advantages of fine-grained features. To mitigate such an issue, a novel regression-based sub-pixel refinement module (SRM) is proposed to exploit the whole fine-grained window features more sufficiently. More specifically, local contexts over windows are aggregated and match information across windows are absorbed firstly. Then both spatial and dimensional similarity between source and target fine-grained window features are gathered to regress fine-level positional deviations. We summarize our contributions in four aspects:

\begin{itemize}
\item A universal FPN-like framework is proposed to capture global structures of not only feature extraction in self-encoder but also match information in cross-decoder by transformers and compensate local contexts as well as implicit positional encoding by convolutions.\smallskip
\item A novel convolutional transformer module is proposed to explore multi-scale long range dependencies between features of multi receptive fields by a novel multi-scale attention and further aggregate local information inside dependencies for enhancing locality by a local pooling unit.\smallskip
\item A novel regression-based sub-pixel refinement module is proposed to aggregate local contexts and match information inside windows then gather both spatial and dimensional similarity between fine-grained window features for fine-level positional deviation regression.\smallskip
\item Experiments on feature matching, homography estimation, relative pose estimation and visual localization benchmarks verify the strong matching capability of LGFCTR.
\end{itemize}

%% file: figure/CNN_TR_CTR.tex
\begin{figure*}[htbp]
  \centering
  \includegraphics[width=0.9\textwidth]{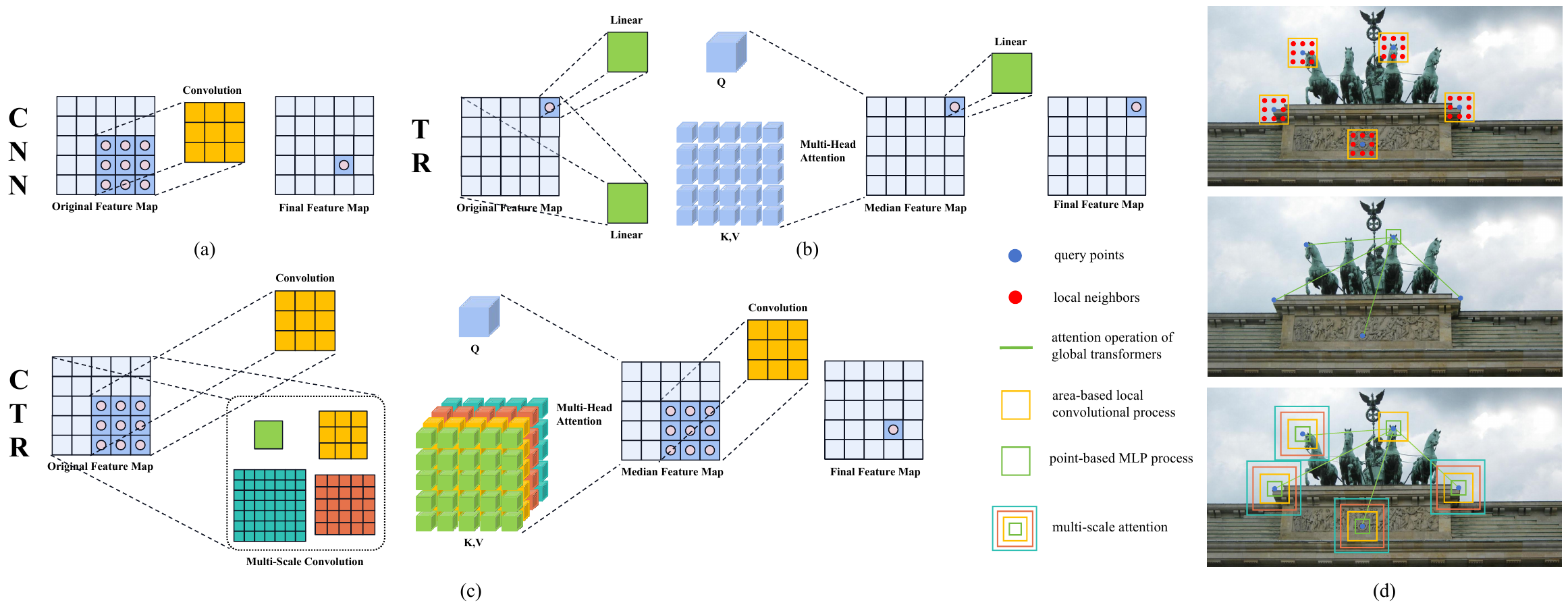}
  \caption{An illustration of comparisons among CNN, transformers (TR) and our convolutional transformer module (CTR). (a) CNN can only extract local contexts due to area-based convolutions. (b) TR can extract global structures through long range attentions but capture few local contexts due to weak locality caused by point-based linear projections. (c) Our CTR can capture both global structures and local contexts more sufficiently through compensating TR for area-based convolutions, where a novel MSA and a LPU are designed. (d) visualizes how the above three modules encode features in different manners.}
  \label{fig:CNN_TR_CTR}
\end{figure*}

%% file: sec/2_related.tex
\section{Related work}
\label{sec:related}

\input{figure/LGFCTR}
\input{figure/CTR}

%-------------------------------------------------------------------------
\subsection{Detector-based image matching}

Before deep learning, handcrafted keypoints and descriptors were designed based on scale-theory by traditional detector-based methods such as SIFT~\cite{sift}, ORB~\cite{orb} and SURF~\cite{surf} and widely used in many downstream tasks due to their efficiency and generalization. Learned detector-based methods follow the same pipeline but use learned features instead to increase the discrimination of keypoints and descriptors. Among them, detect-then-describe methods~\cite{RF-Net, LF-Net, HN++, L2-Net, KeyNet, HAN} extract keypoints~\cite{RF-Net, LF-Net, KeyNet, HAN} and then descriptors~\cite{HN++, L2-Net} based on image patches of certain shapes centered at keypoints, and detect-and-describe methods~\cite{SiLK, SCFeat, DISK, CAPS, GIFT, ASLFeat, R2D2, D2-Net, SP, PoSFeat, Contextdesc} generate both simultaneously. SuperPoint~\cite{SP} as one of the earliest work of the latter type has two parallel branches sharing feature backbone for detection and description. Many other methods~\cite{SiLK, SCFeat, DISK, CAPS, GIFT, ASLFeat, R2D2, D2-Net, Contextdesc} are along this line: D2-Net~\cite{D2-Net} designs how to choose keypoints based on local maximum. R2D2~\cite{R2D2} considers how to measure the reliability of descriptors. ASLFeat~\cite{ASLFeat} fuses DCN operation, peakiness measurement and multi-scale detection together. CAPS~\cite{CAPS} proposes an expectation-based differentiable matching layer and learns robust descriptors using weak epipolar constraints. DISK~\cite{DISK} builds a complete probabilistic model trained by policy gradient. PoSFeat~\cite{PoSFeat} decouples the training of detection and description. Due to the reliance on the repeatability of detection and the reliability of description, they are not robust enough against extreme conditions, including large viewpoint and illumination changes, repetitive patterns, and texture loss.

%-------------------------------------------------------------------------
\subsection{Detector-free image matching}

Detector-free methods eschewed explicit keypoints extraction and generate correspondences directly from raw images in an end-to-end manner, which could take full advantages of rich contexts to achieve more accurate and robust matching than detector-based methods. Before transformers were introduced, NC-Net~\cite{NC-Net} and its variants~\cite{SparseNC-Net, DualRC-Net} generates 4D matching cost volumes followed by 3D convolutions to filter match information, whose receptive field is not enough to capture global features. LoFTR~\cite{LoFTR} and its variants~\cite{AdaMatcher, AspanFormer, COTR, ASTR} adopts transformer-based decoders to capture long range dependencies, which extend the receptive field to the whole image and set state-of-the-art performances. Specifically, SuperGlue~\cite{SG} adopts GNN to process graph nodes constructed by the positions and descriptions of keypoints. LoFTR~\cite{LoFTR} adopts linear transformers~\cite{LinearAttention} to process coarse and fine features extracted by the CNN-based local feature encoder. COTR~\cite{COTR} uses query points to search local features map for target points through a transformer-based decoder. AspanFormer~\cite{AspanFormer} proposes a hierarchical global and local attention framework, where the local attention span is guided by flow maps in a self-adaptive manner. AdaMatcher~\cite{AdaMatcher} performs adaptive assignments on coarse-level matching based on scale estimations between images. ASTR~\cite{ASTR} designs a novel spot-guided transformer-based decoder to avoid interfering with irrelevant areas during feature aggregations. However, these CNN-based encoders still only capture local features. MatchFormer~\cite{MatchFormer} adopts transformer-based encoder to achieve an extract-and-match pipeline and capture long range dependencies but the ability of transformers as decoders is not leveraged completely. Our LGFCTR adopts transformers in both self-encoder and cross-decoder, whose locality is compensated by convolutions. Other dense matching approaches~\cite{GOCor, GLU-Net, DGC-Net, DKM, PDC-Net, IFCAT} focusing on generating dense correspondences over all pixels are also related to our work.

%-------------------------------------------------------------------------
\subsection{Sub-pixel refinement}

Many detector-free methods adopt a coarse-to-fine paradigm for reducing computational cost and guiding fine-level matches by deciding rough areas based on coarse-level matches. Among them, DualRC-Net~\cite{DualRC-Net} finds fine-level targets in the area centered at coarse-level matches and multiplies relevant scores to decide final matches. SparseNC-Net~\cite{SparseNC-Net} operates a softargmax on 3×3 neighborhoods at the soft stage. LoFTR~\cite{LoFTR} according to the expectation-based method in CAPS~\cite{CAPS} regards the spatial expectation of the relevant scores of window features as the sub-pixel matching position. AdaMatcher~\cite{AdaMatcher} takes a further step to process relevant scores with a spatial attention and convolutions. Above methods obtain sub-pixel matches by optimizing matching scores while Patch2Pix~\cite{Patch2Pix} directly regresses relative positional deviations of windows inspired by the bounding box regression in object detection. Our novel SRM belongs to a regression-based method.

%% file: figure/LGFCTR.tex
\begin{figure*}[htbp]
  \centering
  \includegraphics[width=0.9\textwidth]{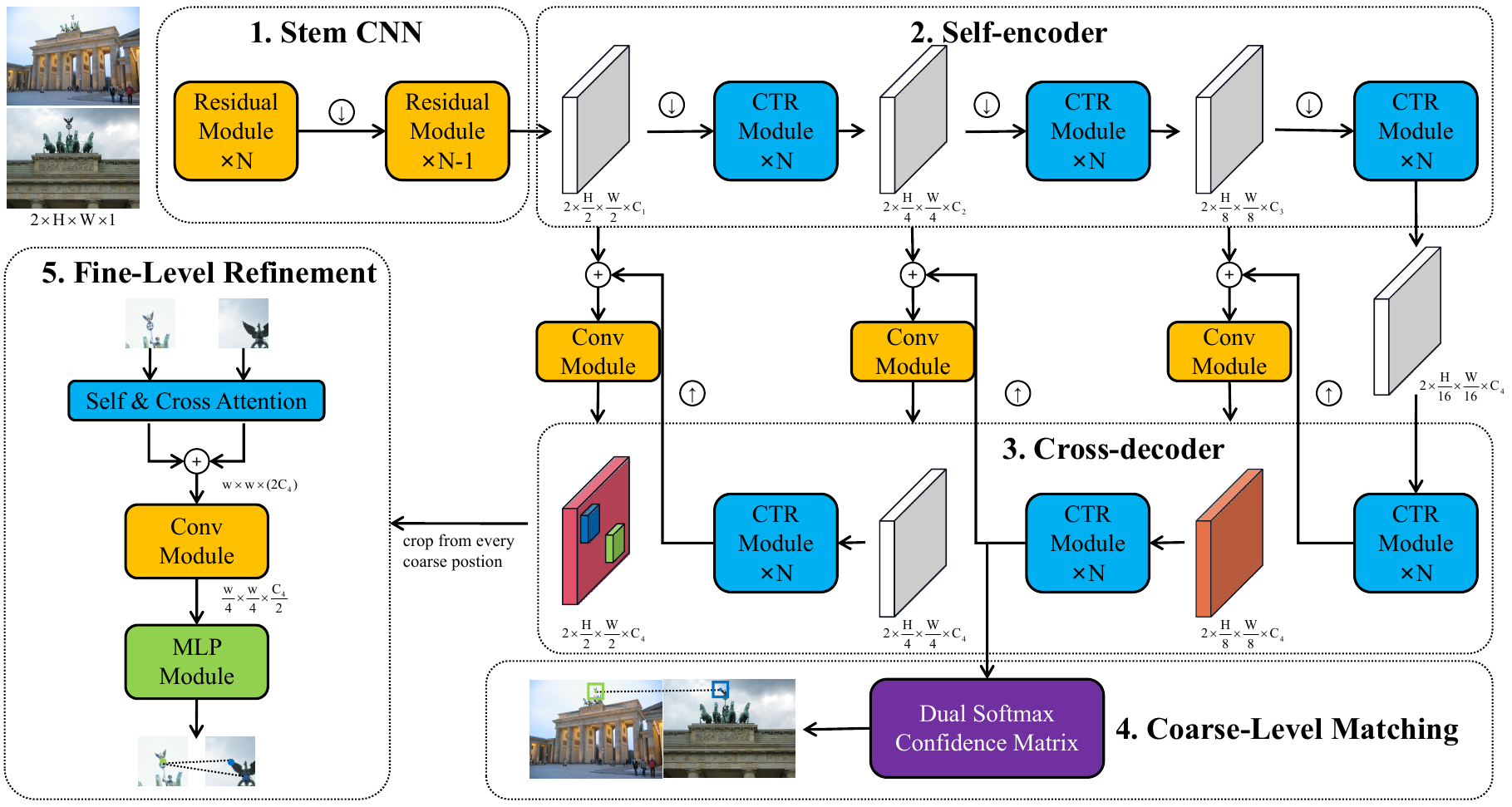}
  \caption{The architecture of LGFCTR. There are five components: 1. A stem CNN extracts shallow local textures and reduces resolution. 2. A transformer-based self-encoder based on CTR capture local and global features in a self manner. 3. A transformer-based cross-decoder does the same thing but in a cross manner. 4. A coarse-level matching module generates a confidence matrix and picks coarse-level matches from it. 5. A fine-level refinement module crops window features at every coarse position and regresses fine-level positional deviations.}
   \label{fig:LGFCTR}
\end{figure*}

%% file: figure/CTR.tex
\begin{figure*}[htbp]
  \centering
  \includegraphics[width=0.9\textwidth]{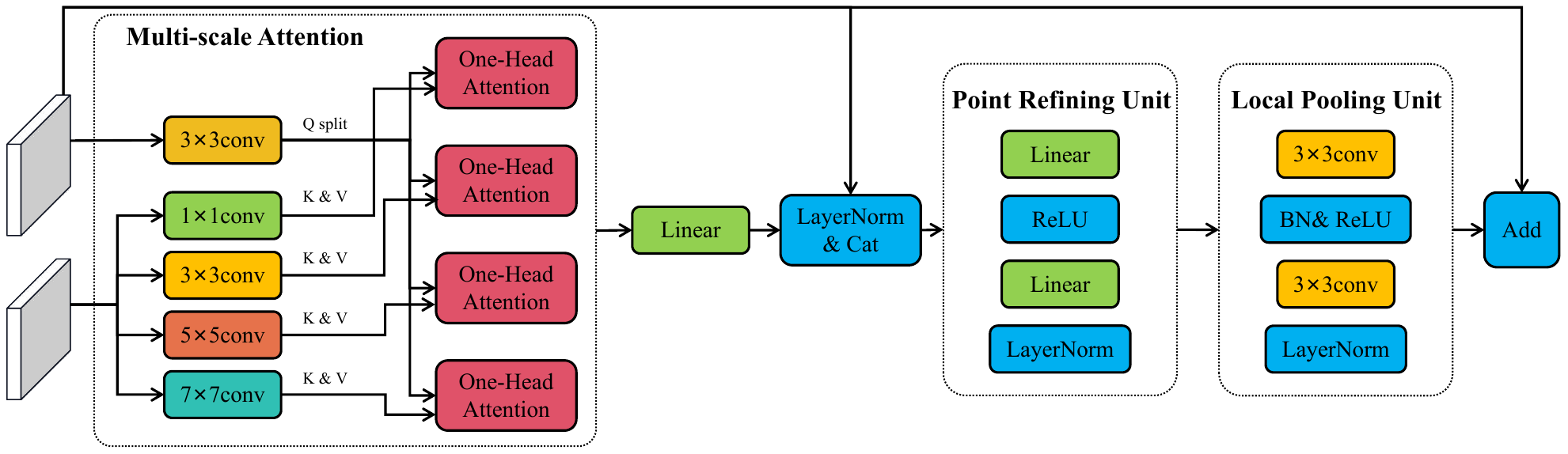}
  \caption{The architecture of the novel convolutional transformer module. In the novel MSA, key and value tokens are generated through convolutions of multi kernel sizes respectively and query tokens are split into four sub-parts along the channel dimension. Then they are respectively fed into each one-head attention. After that, a FFN consisting of a PRU and a LPU further aggregates information in both spatial and channel dimensions. Besides, the positional encoding is achieved by convolutions implicitly.}
  \label{fig:CTR}
\end{figure*}

%% file: sec/3_method.tex
\section{Methodology}
\label{sec:method}

The architecture of LGFCTR is presented in \cref{fig:LGFCTR}. Given a pair of gray images, shallow local textures are extracted and reduced resolution in the stem CNN (\cref{subsec:stem_cnn}). Then shallow features are fed into a FPN-like framework (\cref{subsec:encoder_decoder}) to capture local and global features through a self-encoder and a cross-decoder, where the CTR (\cref{subsec:CTR}) lays the foundation. Among them, coarse-grained features at 1/8 resolution generate a confidence matrix and decide coarse-level matches (\cref{subsec:matches}). Fine-grained features at 1/2 resolution are cropped centered at coarse-level matches and used for regressing fine-level matching adjustments through a SRM (\cref{subsec:matches}).

%-------------------------------------------------------------------------
\subsection{Stem CNN}
\label{subsec:stem_cnn}

Before fed into the transformer-based encoder, a pair of images are processed by a stem CNN first. On one hand, CNN has excellent locality to effectively capture shallow local textures such as edges at high resolution. On the other hand, CNN can reduce the feature resolution efficiently to reduce the computational cost of subsequent transformers.

According to this insight, given a pair of input images $I_A$, $I_B \in R^{H \times W \times 1}$, the stem CNN extracts local features and reduces resolution first, which consists of basic residual modules. Images are processed through $N$ basic residual modules first and then a module whose stride is 2. $F_A^{1/2}$, $F_B^{1/2} \in R^{\frac{H}{2} \times \frac{W}{2} \times C_1}$ are finally given through $N$-1 basic residual modules again.

%-------------------------------------------------------------------------
\subsection{Self-encoder and cross-decoder}
\label{subsec:encoder_decoder}

LGFCTR as a detector-free matcher adopts a universal FPN-like framework. LGFCTR has a transformer-based self-encoder, where the CTR as the basic block compensates transformers for the locality of convolutions. In addition, basic residual modules whose strides are 2 are adopted as feature transitions across different resolutions. Thanks to above two designs, features are encoded with both long range dependencies and local neighborhood constraints. More specifically,
\begin{equation}
\label{eq:1}
\begin{array}{c}
{F_1} = {F_2} = {\rm{downsampling}}(F_j^i)\\
F_j^{i/2} = {\rm{CTR}}({F_1},{F_2}),i \in \{ \frac{1}{2},\frac{1}{4},\frac{1}{8}\} ,j \in \{ A,B\} 
\end{array}
\end{equation}
where the downsampling is achieved by a basic residual module whose stride is 2. Given input features $F_j^{1/2} \in {R^{\frac{H}{2} \times \frac{W}{2} \times {C_1}}}$, multi-scale features $F_j^{1/4} \in {R^{\frac{H}{4} \times \frac{W}{4} \times {C_2}}}$, $F_j^{1/{\rm{8}}} \in {R^{\frac{H}{{\rm{8}}} \times \frac{W}{{\rm{8}}} \times {C_3}}}$, $F_j^{1/{\rm{16}}} \in {R^{\frac{H}{{{\rm{16}}}} \times \frac{W}{{{\rm{16}}}} \times {C_{\rm{4}}}}}$ are obtained through the self-encoder.

Afterwards, a transformer-based cross-decoder adopts multi-scale feature pyramid to compensate up-sampled match features for details with encoder features at each resolution, where match information is decoded in local and global fields within CTRs. More specifically,
\begin{equation}
\label{eq:2}
\begin{array}{c}
{F_1} = {\rm{CNN}}({\rm{upsampling}}(\hat F_j^{i/2}) \oplus F_j^i)\\
{F_2} = {\rm{CNN}}({\rm{upsampling}}(\hat F_{ \sim j}^{i/2}) \oplus F_{ \sim j}^i)\\
\hat F_j^i = {\rm{CTR}}({F_1},{F_2}),i \in \{ \frac{1}{2},\frac{1}{4},\frac{1}{8}\} ,j \in \{ A,B\} 
\end{array}
\end{equation}
where the upsampling is achieved by a bilinear interpolation, $F_{ \sim j}^i$ denotes the other image features, $\oplus$ denotes a concatenation in the channel dimension, ${\rm{CNN}}(\cdot)$ denotes a convolutional neural network, whose details are introduced in the supplementary materials. $\hat F_j^{1/8} \in {R^{\frac{H}{8} \times \frac{W}{8} \times {C_4}}}$, $\hat F_j^{1/4} \in {R^{\frac{H}{4} \times \frac{W}{4} \times {C_4}}}$, $\hat F_j^{1/2} \in {R^{\frac{H}{2} \times \frac{W}{2} \times {C_4}}}$ are obtained through the cross-decoder, among which $\hat F_j^{1/8} \in {R^{\frac{H}{8} \times \frac{W}{8} \times {C_4}}}$ is used for coarse-level matching after flattening into $\hat F_j^{1/8} \in {R^{(\frac{H}{8} \times \frac{W}{8}) \times {C_4}}}$ and $\hat F_j^{1/2} \in {R^{\frac{H}{2} \times \frac{W}{2} \times {C_4}}}$ is used for fine-level matching.

%-------------------------------------------------------------------------
\subsection{Convolutional transformer module}
\label{subsec:CTR}

The novel convolutional transformer module (CTR) is proposed as shown in \cref{fig:CTR}. On one hand, the CTR remains the pattern of summing all spatial features together in an attention manner, which lays the foundation of capturing long range dependencies. On the other hand, the CTR combines area-based convolutions to enhance local context aggregations. The convolution lays on two aspects: 1) Area-based convolutions of multi kernel sizes rather than point-based linear projections are employed in generating tokens of multi receptive fields in the novel multi-scale attention (MSA); 2) The local pooling unit (LPU) as a postprocessing employs a CNN-like structure.

More specifically, given input features ${F_1} \in {R^{H \times W \times C}}$, ${F_{\rm{2}}} \in {R^{H \times W \times C}}$ which are from the same image when in self-encoder and from different images when in cross-decoder, the MSA firstly adopts multi convolutions whose kernel sizes are 1, 3, 5 and 7 respectively to generate four key and value tokens. Compared with point-based linear projections, it can aggregate local contexts in different scales more sufficiently and generate tokens of multi receptive fields. Then query tokens are split into four sub-parts to keep the same channel dimensions with them and fed into four one-head attentions respectively. The attention operations focus on multi-scale long range dependencies between query tokens of fixed receptive field and key tokens of multi receptive fields as shown in \cref{fig:MSA_main}. The above operations can be simplified as:
\begin{equation}
\label{eq:3}
\begin{array}{c}
Q = Con{v_{\rm{3}}}({F_1})\\
K,V = C{v_{\rm{1}}}({F_2}) \oplus C{v_3}({F_2})C{v_5}({F_2}) \oplus C{v_7}({F_2})\\
attn = {\rm{attention}}(Q,K,V)
\end{array}
\end{equation}
where ${\rm{attention}}( \cdot )$ denotes a multi-head linear attention module~\cite{LinearAttention} with four attention heads and $C{v_i}( \cdot )(i = 1,3,5,7)$ denote convolutions for reducing the channels to a quarter of their original sizes, whose kernel sizes are $i$.

Then a post process is employed in aggregating long range dependencies of different scales as:
\begin{equation}
\label{eq:4}
\hat F = {\rm{LN}}(W(attn))
\end{equation}
where ${\rm{LN}}( \cdot )$ denotes a LayerNorm operation and $W( \cdot )$ denotes a linear projection.

Finally, $F_1$ and $\hat{F}$ are concatenated and processed through a Feed Forward Network (FFN) to output $F_1^{'}$ after connecting with $F_1$ in a residual manner. The FFN consists of a point refining unit (PRU) and a LPU. The PRU adopts a standard MLP-like structure to further aggregate multi-scale feature interactions in the channel dimension. The LPU adopts a CNN-like structure to explore local information inside long range dependencies in both spatial and channel dimensions and enhance locality. The above operations can be denoted as:
\begin{equation}
\label{eq:5}
{F'_1} = {F_1} + {\rm{LPU}}({\rm{PRU}}({F_1} \oplus \hat F))
\end{equation}

\input{figure/MSA_main}

%-------------------------------------------------------------------------
\subsection{Matches determination}
\label{subsec:matches}

Following LoFTR~\cite{LoFTR}, LGFCTR adopts a coarse-to-fine paradigm to determine matches.

Coarse-level matches are determined first. The score matrix $S(i,j) = \left\langle {\hat F_A^{1/8}(i),\hat F_B^{1/8}(j)} \right\rangle$ is calculated, where $\left\langle  \cdot  \right\rangle$ denotes a dot-product operation. Then the confidence matrix is generated by employing a dual-softmax operation on $S(i,j)$:
\begin{equation}
\label{eq:6}
{P_c}(i,j) = {\rm{softmax}}{(S(i, \cdot ))_j} \cdot {\rm{softmax}}{(S( \cdot ,j))_i}
\end{equation}
Based on ${P_c}(i,j)$ and the threshold ${\theta _c}$, outliers whose confidences are less than ${\theta _c}$ are filtered and coarse-level matches are determined based on the mutual nearest neighbor criterion. The coarse-level matches are denoted as:
\begin{equation}
\label{eq:7}
{M_c} = \{ (i,j)|\forall (i,j) \in {\rm{MNN}}({P_c}),{P_c}(i,j) \ge {\theta _c}\}
\end{equation}

At the stage of fine-level matching, a novel regression-based sub-pixel refinement module (SRM) is proposed as shown in \cref{fig:LGFCTR}. The fine-grained window features are firstly processed in a self-attention manner followed by a cross-attention operation. It helps aggregate local contexts over windows and absorb match information across windows. Then the SRM stacks two features and gathers both spatial and dimensional similarity between them rather than simple relevant scores. Finally, flattening stacked features into vectors, the relative positional deviations of windows are obtained in a regression manner, which exploits the whole fine-grained window features more sufficiently rather than only the central feature of source window. Compared to expectation-based methods, our SRM achieves more accurate fine-level matching as shown in \cref{fig:subpixel_main}. More specifically, two input fine-grained window features $\hat F_{A,i}^{1/2}$, $\hat F_{B,j}^{1/2} \in {R^{w \times w \times {C_4}}}$ are cropped from $\hat F_j^{1/2} \in {R^{\frac{H}{2} \times \frac{W}{2} \times {C_4}}}$ at $(i,j) \in {M_c}$ and then processed by:
\begin{equation}
\label{eq:8}
\begin{array}{c}
\tilde F_{A,i}^{1/2},\tilde F_{B,j}^{1/2} = {\rm{LoFTR}}(\hat F_{A,i}^{1/2},\hat F_{B,j}^{1/2})\\
(\Delta x,\Delta y) = {\rm{MLP}}({\rm{CNN}}(\tilde F_{A,i}^{1/2} \oplus \tilde F_{B,j}^{1/2}))
\end{array}
\end{equation}
where ${\rm{CNN}}( \cdot )$ shares the same structure as the CNN mentioned in \cref{subsec:encoder_decoder}, ${\rm{MLP}}( \cdot )$ denotes a multiple layer perceptron, $(\Delta x,\Delta y)$ denotes a relative deviation of the window, and ${\rm{LoFTR}}( \cdot )$ denotes a fine-level transformer module proposed by LoFTR~\cite{LoFTR} which contains a self-attention and a cross-attention. The details are introduced in the supplementary materials. After the sub-pixel adjustment, fine-level matches ${M_f}{\rm{ = \{ (}}i,\hat j{\rm{)\} }}$ are determined.

\input{figure/subpixel_main}

%-------------------------------------------------------------------------
\subsection{Loss formulation}
\label{subsec:loss}

Following LoFTR~\cite{LoFTR}, our loss function consists of a coarse-level loss and a fine-level loss, which can be denoted as $L=L_c+L_f$. The groundtruth matches are derived according to depth and camera poses. For coarse-level loss, we use the same Focal Loss~\cite{FocalLoss} as in LoFTR, where $\alpha$ and $\beta$ of Focal Loss are set to 0.25 and 2, respectively. For fine-level loss, we minimize the L2 distances between fine-level matches $M_f$ and groundtruth $M_f^{gt}$ as in LoFTR but remove uncertainty measurement:
\begin{equation}
\label{eq:9}
{L_f} = \frac{1}{{\left| {{M_f}} \right|}}\sum\limits_{(i,\hat j) \in {M_f}} {{{\left\| {\hat j - {{\hat j}_{gt}}} \right\|}_2}}
\end{equation}

%-------------------------------------------------------------------------
\subsection{Implementation details}
\label{subsec:implementation}

We train LGFCTR on the MegaDepth~\cite{MegaDepth} dataset following LoFTR~\cite{LoFTR}. The model is trained using an Adamw optimizer with an initial learning rate of 5e-4, a batch size of 4 and a weight decay of 0.1 for 30 epochs on 4 GTX 3090 GPUs. In the training, 100 pairs of images are sampled from each sub-scene and resized with longer dimensions equal to 800. The number $N$ of residual modules and CTRs is both 3. The window size in fine-level matching stage is equal to 5. $\theta_c$ is set to 0.2. $C_1$, $C_2$, $C_3$, $C_4$ are set to 64, 128, 192, 256, respectively. More details are introduced in the supplementary materials.

%% file: figure/MSA_main.tex
\begin{figure*}[htbp]
  \centering
  \includegraphics[width=0.9\textwidth]{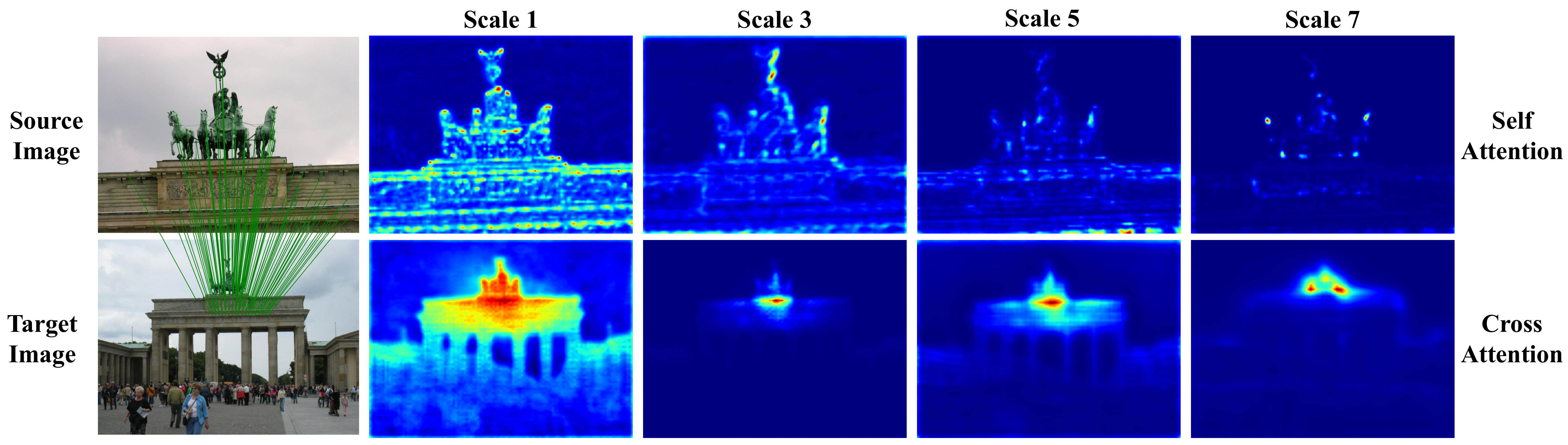}
  \caption{The visualizations of multi-scale attention. MSA of the last CTR at 1/8 resolution are chosen. The query is the center of source image and the key is the whole source image (self-attention) and target image (cross-attention), respectively. Warmer colors represent higher attention weights. The MSA captures different texture features in different scales and puts the focuses around different target areas. The multi-scale long range dependencies are gathered together to achieve more accurate and comprehensive attention patterns.}
  \label{fig:MSA_main}
\end{figure*}

%% file: figure/subpixel_main.tex
\begin{figure}[htbp]
  \centering
  \includegraphics[width=0.9\linewidth]{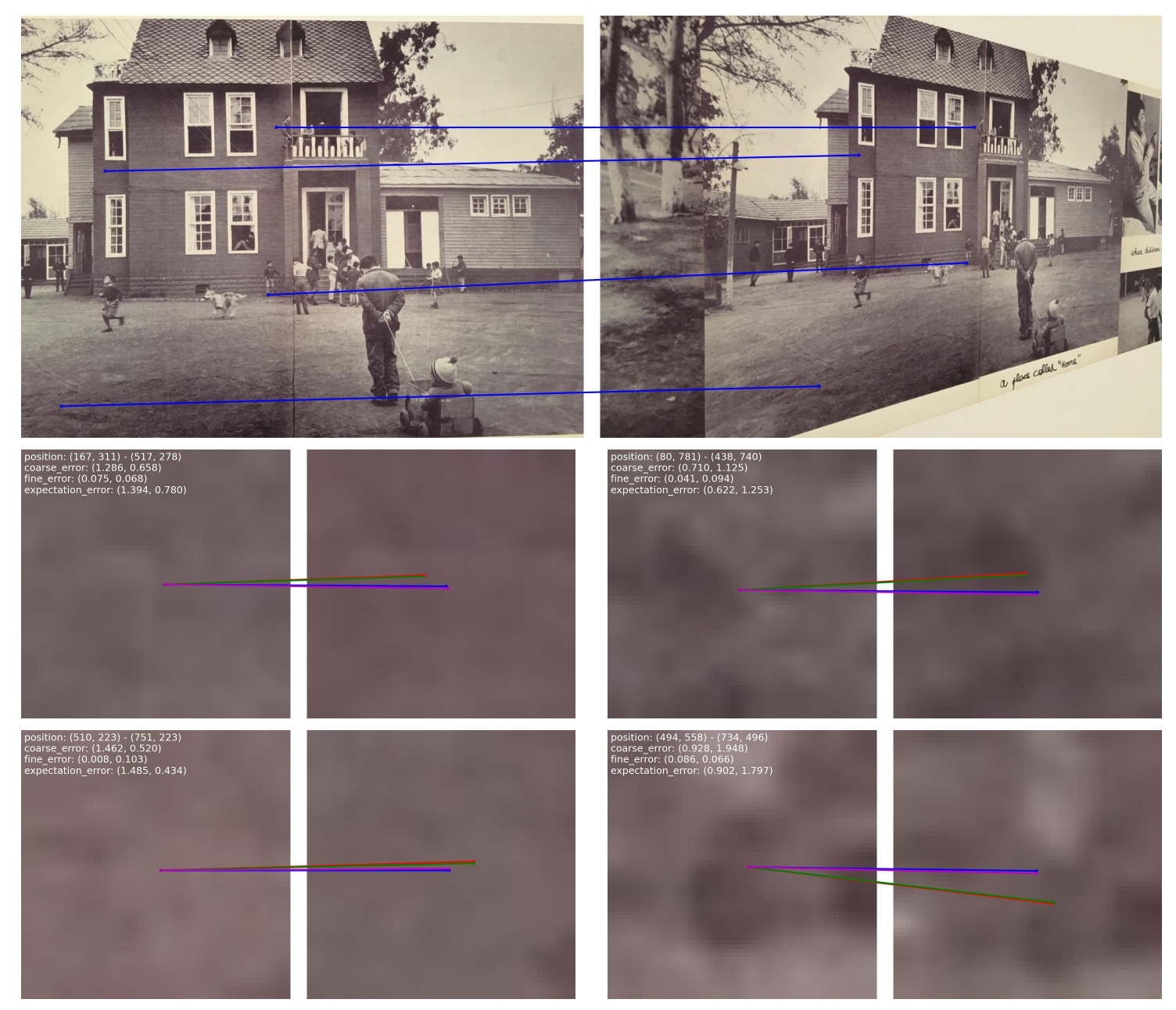}
  \caption{The visualizations of sub-pixel refinement module. The top row shows four matches across a pair of images from HPatches dataset. The two bottom rows shows the zoom-in patches and matching details. The blue, red, green and purple lines respectively represent coarse matches, groundtruths, fine matches with SRM and fine matches with expectations. The positions and matching errors are annotated on the top left of each pair of patches.}
  \label{fig:subpixel_main}
\end{figure}

%% file: sec/4_experiments.tex
\section{Experiments}
\label{sec:experiments}

%-------------------------------------------------------------------------
\subsection{Homography estimation}
\label{subsec:homo}

\noindent \textbf{Datasets.} We firstly evaluate LGFCTR on HPatches\cite{HPatches} dataset for homography estimation. HPatches contains 52 sequences under significant illumination changes and 56 sequences under obvious viewpoint changes. There are 1 reference image and 5 query images in each sequence.

\noindent \textbf{Evaluation protocol.} All images are resized with longer dimensions equal to 640. All matches are used for homography estimation with RANSAC, which is achieved by OpenCV. We compute the corner error between the images warped with the estimated $\hat{H}$ and the groundtruth $H$ following ~\cite{LoFTR} and report the accuracy at the threshold of 3 pixels.

\noindent \textbf{Results.} As shown in \cref{tab:homo}, LGFCTR outperforms all comparative methods with a margin of as least 1.7\% in the overall sequences. Though DualRC-Net~\cite{DualRC-Net} achieves an advantage in the illumination sequences, it is not robust enough against large viewpoint changes. SP~\cite{SP} + SG~\cite{SG} achieves excellent performances in the viewpoint sequences but is inferior to LGFCTR in the illumination sequences with a large margin of 5.4\%.

\input{table/homo_acc}

%-------------------------------------------------------------------------
\subsection{Relative pose estimation}
\label{subsec:pose}

\noindent \textbf{Datasets.} We evaluate LGFCTR on MegaDepth~\cite{MegaDepth} and YFCC100M~\cite{YFCC100M} datasets for relative pose estimation. Same as other works~\cite{LoFTR, AspanFormer}, we make evaluations on 1500 pairs sampled from MegaDepth and 4000 pairs sampled from YFCC100M.

\noindent \textbf{Evaluation protocol.} All images of MegaDepth are resized with longer dimensions equal to 1200 for all methods except 1216 for AspanFormer~\cite{AspanFormer} and 1152 for AdaMatcher~\cite{AdaMatcher}. We use OpenCV to compute the essential matrix with RANSAC and then recover the rotation and translation matrix. In terms of YFCC100M, the RANSAC threshold is set to 5e-4 in the normalized coordinate space for all methods following~\cite{AspanFormer}. The pose error is defined as the maximum of angular errors in rotation and translation. We report the area under the cumulative curve (AUC) of pose errors at the thresholds of 5°, 10°, 20°, respectively.

\noindent \textbf{Results.} As shown in \cref{tab:pose_MegaDepth}, LGFCTR outperforms all comparative methods on MegaDepth dataset with a margin of 2.29\% in AUC@5°, 1.71\% in AUC@10°, 0.74\% in AUC@20°. The qualitative results are shown in \cref{fig:MegaDepth_main}. With respect to results on YFCC100M dataset as shown in \cref{tab:pose_YFCC100M}, LGFCTR improves by 0.4\%, 0.3\%, 0.2\% respectively at the thresholds of 5°, 10°, 20° against all comparative methods.

\input{figure/MegaDepth_main}
\input{table/pose}

%-------------------------------------------------------------------------
\subsection{Visual localization}
\label{subsec:loc}

\noindent \textbf{Datasets.} We make an evaluation on Aachen Day-Night v1.1~\cite{Aachen} dataset for visual localization. Aachen Day-Night v1.1 dataset depicts a city whose reference scene model is built upon 6,697 day-time images. The dataset provides 824 day-time images and 191 night-time images as queries. For each query image, we need to generate multi-view correspondences with reference images and compute a 6DoF pose based on known poses of reference images.

\noindent \textbf{Evaluation protocol.} We use localization pipeline HLoc~\cite{HLoc} to compute query poses and obtain quantitative results through Long-Term Visual Localization Benchmark~\cite{loc_benchmark}. We report the accuracy at the thresholds of (0.25m, 2°), (0.5m, 5°), (1.0m, 10°), respectively.

\noindent \textbf{Results.} In overall, LGFCTR is on par with state-of-the-art methods and achieves comparative performances in both day-time and night-time scenes. It proves the strong generalization of LGFCTR on visual localization.

\input{table/loc}

%-------------------------------------------------------------------------
\subsection{Ablation study}
\label{subsec:ablation}

To improve efficiency, we trained 6 lite models on MegaDepth dataset with a different setting to study the effectiveness of each component including positional encoding (PE), SRM, LPU and MSA, whose details are described in the supplementary materials. The evaluations on MegaDepth dataset following \cref{subsec:pose} are listed in \cref{tab:ablation}.

\noindent \textbf{Effectiveness of SRM}. Index-1 improves by 0.46\%, 1.07\%, 0.77\% respectively against Index-2. It proves the novel SRM can adjust more accurate fine-level matches.

\noindent \textbf{Effectiveness of MSA}. Index-1 improves by 0.10\%, 0.40\%, 0.16\% respectively against Index-3 and Index-4 improves by 0.45\%, 0.38\%, 0.45\% respectively against Index-5. It proves the novel MSA can capture global structures more sufficiently to generate more valid matches.

\noindent \textbf{Effectiveness of LPU}. Index-1 improves by 1.06\%, 1.68\%, 1.14\% respectively against Index-4 and Index-3 improves by 1.41\%, 1.28\%, 0.98\% respectively against Index-5. It proves the LPU can further aggregate local information to enhance locality of matching.

\noindent \textbf{Effectiveness of PE}. Index-5 improves by 2.85\%, 3.01\%, 2.87\% respectively against Index-6. It proves the convolution contributes enough to implicit positional encoding and PE hurt the generalization on different testing resolutions. The impact of input resolution on PE are discussed in the supplementary materials.

\input{table/ablation}

%% file: table/homo_acc.tex
\begin{table}[htbp]
\centering
\resizebox{\linewidth}{!}
{
\begin{tabular}{@{}ccccc@{}}
\toprule
\multirow{2}{*}{Category}       & \multirow{2}{*}{Method} & Illumination            & Viewpoint               & Overall                 \\ \cmidrule(l){3-5} 
                                &                         & \multicolumn{3}{c}{ACC@3px}                                      \\ \midrule
\multirow{5}{*}{Detector-based} & D2-Net\cite{D2-Net}            & 66.9    & 13.6     & 39.3    \\
                                & R2D2\cite{R2D2}                & 90.8    & 42.5     & 65.7    \\
                                & DISK\cite{DISK}                & 89.2    & 46.4     & 67.0    \\
                                & SIFT\cite{sift}+ CAPS\cite{CAPS}       & 85.8      & 48.6       & 66.5    \\
                                & SP\cite{SP} + SG\cite{SG}      & 89.6    & \textbf{53.0}     & 70.7    \\ \midrule
\multirow{5}{*}{Detector-free}  & DualRC-Net\cite{DualRC-Net}    & \textbf{96.5}    & 8.6      & 50.9    \\
                                & AdaMatcher\cite{AdaMatcher}    & 78.5    & 43.8     & 60.2    \\
                                & MatchFormer\cite{MatchFormer}  & 94.6    & 46.4     & 69.6    \\
                                & LoFTR\cite{LoFTR}              & 94.6    & 48.9     & {\ul70.9}    \\
                                & LGFCTR                         & {\ul95.0}    & {\ul51.8}     & \textbf{72.6}    \\ \bottomrule
\end{tabular}
}
\caption{Homography estimation on HPatches dataset.}
\label{tab:homo}
\end{table}

%% file: figure/MegaDepth_main.tex
\begin{figure*}[htbp]
  \centering
  \includegraphics[width=0.9\textwidth]{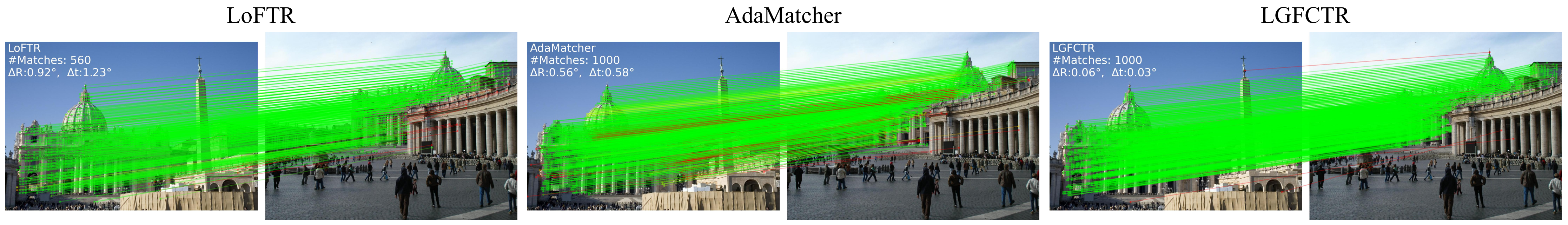}
  \caption{The qualitative results on MegaDepth dataset. The matches shown and estimated pose errors are annotated on the top left of each pair of images.}
  \label{fig:MegaDepth_main}
\end{figure*}

%% file: table/pose.tex
\begin{table}[htbp]
\centering
\footnotesize
\begin{tabular}{@{}ccccc@{}}
\toprule
\multirow{2}{*}{Category}       & \multirow{2}{*}{Method}        & \multicolumn{3}{c}{Pose estimation AUC}          \\ \cmidrule(l){3-5} 
                                &                                & @ 5°           & @ 10°          & @ 20°          \\ \midrule
\multirow{5}{*}{Detector-based} & SP\cite{SP} + SG\cite{SG} & 42.18    & 61.16   & 75.96    \\
                                & D2-Net\cite{D2-Net}                     & 19.60           & 36.33          & 54.59          \\
                                & R2D2\cite{R2D2}                       & 36.29          & 54.06          & 69.41          \\
                                & SIFT\cite{sift} + CAPS\cite{CAPS}             & 29.24          & 45.28          & 60.90           \\
                                & DISK\cite{DISK}                       & 38.63          & 54.92          & 68.00             \\ \midrule
\multirow{8}{*}{Detector-free}  & LoFTR-DS\cite{LoFTR}                   & 52.80           & 69.19          & 81.18          \\
                                & MatchFormer\cite{MatchFormer}                & 52.91          & 69.74          & 82.00             \\
                                & AspanFormer\cite{AspanFormer}                & 55.32          & 71.73          & 83.07          \\
                                & AdaMatcher\cite{AdaMatcher}                 & 52.46           & 69.78          & 82.19          \\
                                & ASTR\cite{ASTR}                             & {\ul 58.40}           &
                                {\ul 73.10}          & 83.80          \\
                                & SEM\cite{SEM}                               & 58.00           &
                                72.90          & 83.70          \\
                                & PATS\cite{PATS}                             & 56.92           &
                                73.04          & {\ul 84.06}          \\
                                & LGFCTR                         & \textbf{60.69} & \textbf{74.81} & \textbf{84.80} \\ \bottomrule
\end{tabular}
\caption{Relative pose estimation on MegaDepth dataset.}
\label{tab:pose_MegaDepth}
\end{table}

\begin{table}[htbp]
\centering
\resizebox{\linewidth}{!}
{
\begin{tabular}{@{}ccccc@{}}
\toprule
\multirow{2}{*}{Category}       & \multirow{2}{*}{Method}        & \multicolumn{3}{c}{Pose estimation AUC}          \\ \cmidrule(l){3-5} 
                                &                                & @ 5°           & @ 10°          & @ 20°          \\ \midrule
\multirow{2}{*}{Detector-based} & SP\cite{SP} + SG\cite{SG} & 38.7    & 59.1    & 75.8    \\
                                & RootSIFT\cite{rootsift} + SGMNet\cite{SGMNet}                       & 35.5          & 55.2          & 71.9             \\ \midrule
\multirow{6}{*}{Detector-free}  & DualRC-Net\cite{DualRC-Net}                   & 29.5           & 50.1          & 66.8          \\
                                & LoFTR\cite{LoFTR}                & 42.4          & 62.5          & 77.3             \\
                                & AspanFormer\cite{AspanFormer}                & 44.5          & 63.8          & 78.4          \\
                                & AdaMatcher\cite{AdaMatcher}                 & 29.2           & 45.2          & 59.5          \\
                                & PATS\cite{PATS}                             & {\ul 45.8}           &
                                {\ul 64.7}          & {\ul 78.9}          \\
                                & LGFCTR                         & \textbf{46.2}               & \textbf{65.0}          & \textbf{79.1} \\ \bottomrule
\end{tabular}
}
\caption{Relative pose estimation on YFCC100M dataset.}
\label{tab:pose_YFCC100M}
\end{table}

%% file: table/loc.tex
\begin{table}[htbp]
\centering
\resizebox{\linewidth}{!}
{
\begin{tabular}{@{}ccccc@{}}
\toprule
\multirow{2}{*}{Category}       & \multirow{2}{*}{Method} & Day                          & \multicolumn{2}{c}{Night}                       \\ \cmidrule(l){3-5} 
                                &                         & \multicolumn{3}{c}{(0.25m, 2°) / (0.5m, 5°) / (1.0m, 10°)}                     \\ \midrule
\multirow{4}{*}{Detector-based} & D2-Net\cite{D2-Net}              & 84.5 / 91.3 / 96.1           & \multicolumn{2}{c}{67.0 / 80.6 / 90.6}          \\
                                & R2D2\cite{R2D2}                & 89.7 / 95.8 / 98.8           & \multicolumn{2}{c}{68.1 / 84.8 / 93.7}          \\
                                & DISK\cite{DISK}                & 86.5 / 94.3 / 98.1           & \multicolumn{2}{c}{72.3 / 83.8 / 93.7}          \\
                                & SP\cite{SP} + SG\cite{SG}         & {\ul 89.8} / {\textbf{96.1}} / {\textbf{99.4}}  & \multicolumn{2}{c}{77.0 / 90.6 / {\textbf{100.0}}}         \\ \midrule
\multirow{6}{*}{Detector-free}  & LoFTR\cite{LoFTR}               & 88.7 / 95.6 / 99.0           & \multicolumn{2}{c}{{\ul 78.5} / 90.6 / 99.0}          \\
                                & AspanFormer\cite{AspanFormer}         & 89.4 / 95.6 / 99.0           & \multicolumn{2}{c}{77.5 / {\ul 91.6} / {\ul 99.5}}          \\
                                & AdaMatcher\cite{AdaMatcher}    & 89.2 / {\ul 96.0} / {\ul 99.3}           & \multicolumn{2}{c}{{\textbf{79.1}} / 90.6 / {\ul 99.5}} \\
                                & ASTR\cite{ASTR}    & {\textbf{89.9}} / 95.6 / 99.2           & \multicolumn{2}{c}{76.4 / {\textbf{92.1}} / {\ul 99.5}} \\
                                & PATS\cite{PATS}    & 89.6 / 95.8 / {\ul 99.3}           & \multicolumn{2}{c}{73.8 / {\textbf{92.1}} / {\ul 99.5}} \\
                                & LGFCTR                  & {\ul89.8} / 95.9 / 98.9           & \multicolumn{2}{c}{75.9 / 90.1 / {\ul 99.5}}          \\ \bottomrule
\end{tabular}
}
\caption{Visual localization on Aachen Day-Night v1.1 dataset.}
\label{tab:loc}
\end{table}

%% file: table/ablation.tex
\begin{table}[htbp]
\centering
\footnotesize
\begin{tabular}{@{}cccccccc@{}}
\toprule
\multirow{2}{*}{Index} & \multirow{2}{*}{PE}  & \multirow{2}{*}{SRM} & \multirow{2}{*}{LPU} & \multirow{2}{*}{MSA} & \multicolumn{3}{c}{Pose estimation AUC}                            \\ \cmidrule(l){6-8} 
                       &                      &                      &                      &                      & @ 5°                 & @ 10°                & @ 20°                \\ \midrule
1                      &                      & \checkmark                    & \checkmark                    & \checkmark                    & \textbf{55.55}                & \textbf{71.85}                & \textbf{83.10}                \\
2                      &                      &                      & \checkmark                    & \checkmark                    & 55.09                      & 70.78                     & 82.33                     \\
3                      &                      & \checkmark                    & \checkmark                    &                      & 55.45                     & 71.45                     & 82.94                     \\
4                      &                      & \checkmark                    &                      & \checkmark                    & 54.49                     & 70.55                     & 82.41                     \\
5                      &                      & \checkmark                     &                      &                      & 54.04                     & 70.17                    & 81.96                     \\
6                      & \checkmark                    & \checkmark                     &                      &                      & 51.19                     & 67.16                     & 79.09                     \\
\bottomrule
\end{tabular}
\caption{The ablation study of each component.}
\label{tab:ablation}
\end{table}

%% file: sec/5_conclusion.tex
\section{Conclusion}
\label{sec:conclusion}

In this paper, we propose a novel convolutional transformer to capture both global structures and local contexts more sufficiently for detector-free matching, including a universal FPN-like framework fusing transformers and convolutions in both self-encoder and cross-decoder, a novel convolutional transformer module exploring multi-scale long range dependencies and enhancing locality, and a novel regression-based sub-pixel refinement module exploiting the whole fine-grained window features more sufficiently. LGFCTR is on par with or outperforms state-of-the-art methods on a wide range of benchmarks, including feature matching, homography estimation, relative pose estimation and visual localization. We are hopeful that such a convolutional transformer framework will bring a new insight to image matching community and show stronger capability with more computational resources.

\noindent \textbf{Acknowledgement}. This research was supported by the National Natural Science Foundation of China (NSFC) under grant number 61725501.

%% file: sec/X_suppl.tex
\clearpage
\setcounter{page}{1}
\setcounter{section}{0}
\setcounter{figure}{0}
\setcounter{table}{0}
\maketitlesupplementary

%-------------------------------------------------------------------------

\section{Implementation details}
\label{sec:supplementary_implementation}

%-------------------------------------------------------------------------
\subsection{Architecture of sub-pixel refinement module}
\label{subsec:sub-pixel_refinement_module}

The architecture of CNN and MLP of sub-pixel refinement module (SRM) in Section 3.4 of the main text is shown in \cref{fig:CNN_Res}(b) and (c), respectively. The CNN of sub-pixel refinement module shares the same structure as \cref{fig:CNN_Res}(a) but reduce features resolutions and channels to one half of the origin sizes in both two convolution layers. The MLP firstly adopts two sub-blocks consisting of a fully connected layer and a leaky rectified linear unit, which reduce features channels to a quarter of the origin sizes. Then the features are projected to two channels and normalized to the range from -1 to 1, corresponding to normalized positional deviations in X and Y axis respectively.

The architecture of self and cross attention of SRM, which is the same as fine-level transformer module proposed by LoFTR~\cite{LoFTR}, is shown in \cref{fig:LoFTR}. The multi-head attention has 8 heads and the MLP-like feed forward network is the same as point refining unit (PRU), whose details are provided in \cref{subsec:other_modules}.

%-------------------------------------------------------------------------
\subsection{Details about several other modules}
\label{subsec:other_modules}

The architecture of CNN of Eq. (2) in Section 3.2 of the main text is shown in \cref{fig:CNN_Res}(a). It is a sequential process consisting of a convolution layer which outputs target features channels, a batch normalization, a rectified linear unit and a convolution layer keeping the feature channel. It helps aggregate semantic information of up-sampled match features and texture details of encoder features at the same resolution.

A PRU consists of two fully connected layers, a rectified linear unit and a layer normalization. The first fully connected layer keeps features channels and the second reduce channels to one half of the origin sizes.

A local pooling unit (LPU) consists of two convolution layers, a batch normalization, a rectified linear unit and a layer normalization. Both convolution layers keep features channels.

The basic residual block adopts the same structure as LoFTR~\cite{LoFTR}, which is the basic block proposed by ResNet~\cite{ResNet}.

\input{figure/CNN_Res}

%-------------------------------------------------------------------------
\subsection{Training details}
\label{subsec:training_details}

The training of LGFCTR mostly followed LoFTR~\cite{LoFTR}. To be detailed, the learning rate is increasing linearly from 5e-5 to 5e-4 within the first three epochs and decays in 8, 12, 16, 20, 24 epochs with a rate of 0.5.
In the ablation study, a different setting is chosen to improve efficiency. To be detailed, $C_1$, $C_2$, $C_3$, $C_4$ are set to 64, 64, 96, 128, respectively. In the training, each model was trained with only 30 pairs of images sampled from each sub-scene on 2 GTX 3090 GPUs.

%-------------------------------------------------------------------------
\section{More experiments and details}
\label{sec:supplementary_experiments}

%-------------------------------------------------------------------------
\subsection{The impact of input resolution on positonal encoding}
\label{subsec:feature_matching}

This part extends the analysis of effectiveness of positional encoding from Section 4.4 of the main text. Index-1, Index-5 and Index-6 are further evaluated with input resolutions of 600, 800, 1000 and the results are listed in \cref{tab:input_resolution}.

Without explicit positional encoding, Index-1 and Index-5 benefit from more details brought from increasing input resolution and achieve higher quantitative results of AUC at all thresholds. More specifically, for example, Index-1 improves by 4.15\%, 1.67\%, 1.21\% in AUC@10° as the input resolution increases progressively from 600 to 1200, while Index-5 improves by 4.59\%, 1.52\%, 1.40\%. However, while inserting explicit positional encoding, Index-6 achieves a performance gain until the input resolution increases to 1000 and decays in the subsequent resolution. More specifically, Index-6 improves by 6.73\%, 6.79\%, 6.71\% while the input resolution increases from 600 to 800 but decays by 0.12\%, 0.67\%, 1.31\% while the input resolution increases from 1000 to 1200, and achieves best performances around the input resolution of 1000, which is close to the training resolution. It indicates that explicit positional encoding will lead transformers to fit into the training resolution. It further proves the convolution contributes enough to implicit positional encoding and explicit positional encoding will hurt the generalization on different testing resolutions.

In addition, \cref{tab:input_resolution} shows that Index-1 outperforms Index-5 in all input resolutions, which further proves the effectiveness of convolutional transformer module (CTR), multi-scale attention (MSA) and LPU.

\input{table/input_resolution}

%-------------------------------------------------------------------------
\subsection{Feature matching}
\label{subsec:feature_matching}

\noindent \textbf{Evaluation protocol}. We provide another evaluation on HPatches~\cite{HPatches} dataset for feature matching. Same as Section 4.1 of the main text, all images are resized with longer dimensions equal to 640. The mean matching accuracy (MMA) at thresholds ranging from 1 to 10 pixels are reported as quantitative results. In addition, in order for overall evaluation across different thresholds, the MMA scores are also reported following PoSFeat~\cite{PoSFeat}.

\noindent \textbf{Results}. The MMA results at different thresholds are shown in \cref{fig:feature_matching} and the MMA scores are listed in \cref{tab:feature_matching}. In the overall sequences, LGFCTR outperforms all comparative methods up to 4px and achieves the highest MMA score of 0.8156, which has a large margin of 0.0239 against the second place. In the illumination sequences, DualRC-Net~\cite{DualRC-Net} achieves the best performances but is not robust enough against large viewpoint changes. Actually, LGFCTR has similar performances to DualRC-Net while threshold is larger than 5px. In terms of viewpoint sequences, LGFCTR is superior to all comparative methods except for SP~\cite{SP} + SG~\cite{SG}. However, LGFCTR outperforms SP + SG up to 3px in the viewpoint sequences and is superior with a large margin of 0.0369 in the overall sequences.

\input{figure/LoFTR}
\input{figure/feature_matching}
\input{table/feature_matching}

%-------------------------------------------------------------------------
\subsection{Relative pose estimation}
\label{subsec:pose_supple}

We also evaluate DKM~\cite{DKM} on MegaDepth dataset with inner resized resolution of 1200 in width and 900 in height to keep the same as other methods, while its original inner resized resolution is 880 in width an 660 in height. DKM achieves performances of 58.40\%, 73.41\% and 84.10\% at the thresholds of 5°, 10° and 20°, while our method LGFCTR outperform it with the results of 60.69\%, 74.81\% and 84.80\%. Different from LGFCTR, DKM belongs to a dense matching method which predict correspondence of every pixel at query image. So it is not considered as a comparative method in the main text.

%-------------------------------------------------------------------------
\section{Qualitative results}
\label{sec:qualitative}

More qualitative results and visualizations are provided in this part. The qualitative comparisons among LoFTR~\cite{LoFTR}, AdaMatcher~\cite{AdaMatcher} and LGFCTR on illumination and viewpoint sequences of HPatches dataset~\cite{HPatches} are provided in \cref{fig:HPatches_i} and \cref{fig:HPatches_v}, respectively. Besides, their more qualitative results on MegaDepth dataset~\cite{MegaDepth} are also provided in \cref{fig:MegaDepth_suppl}. It can be seen that LGFCTR achieves more accurate matching and relative pose estimations than other comparative methods. More visualizations of SRM are provided in \cref{fig:subpixel_suppl}. More visualizations of MSA are provided in \cref{fig:subpixel_suppl}

\input{figure/HPatches_i}
\input{figure/HPatches_v}
\input{figure/MegaDepth_suppl}
\input{figure/subpixel_suppl}
\input{figure/MSA_suppl}

%% file: figure/CNN_Res.tex
\begin{figure}[htbp]
  \centering
  \includegraphics[width=0.95\linewidth]{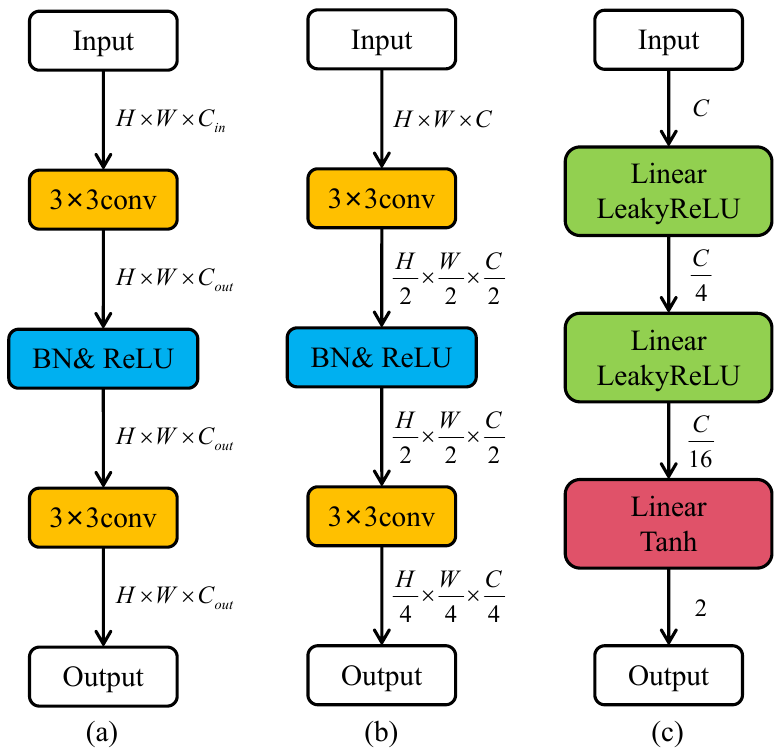}
  \caption{The architecture of several modules proposed in the main text. (a) is the architecture of CNN in Section 3.2 of the main text; (b) and (c) are the architecture of CNN and MLP in the sub-pixel refinement module.}
  \label{fig:CNN_Res}
\end{figure}

%% file: table/input_resolution.tex
\begin{table}[htbp]
\centering
\resizebox{\linewidth}{!}
{
\begin{tabular}{@{}cccc@{}}
\toprule
\multirow{2}{*}{Input resolution} & \multicolumn{3}{c}{Pose estimation AUC@10°} \\ \cmidrule(l){2-4} 
                                  & Index-1       & Index-5      & Index-6      \\ \midrule
600                               & 47.93 / 64.82 / 77.98             & 46.30 / 62.66 / 75.72            & 44.28 / 60.48 / 72.93            \\
800                               & 52.66 / 68.97 / 81.03             & 50.93 / 67.25 / 79.56            & 51.01 / 67.27 / 79.64            \\
1000                              & 54.26 / 70.64 / 82.47             & 52.77 / 68.77 / 71.01            & 51.31 / 67.83 / 80.40            \\ 
1200                              & 55.55 / 71.85 / 83.10             & 54.04 / 70.17 / 81.96            & 51.19 / 67.16 / 79.09            \\ \bottomrule
\end{tabular}
}
\caption{The impact of input resolution on positional encoding.}
\label{tab:input_resolution}
\end{table}

%% file: figure/LoFTR.tex
\begin{figure}[htbp]
  \centering
  \includegraphics[width=1.0\linewidth]{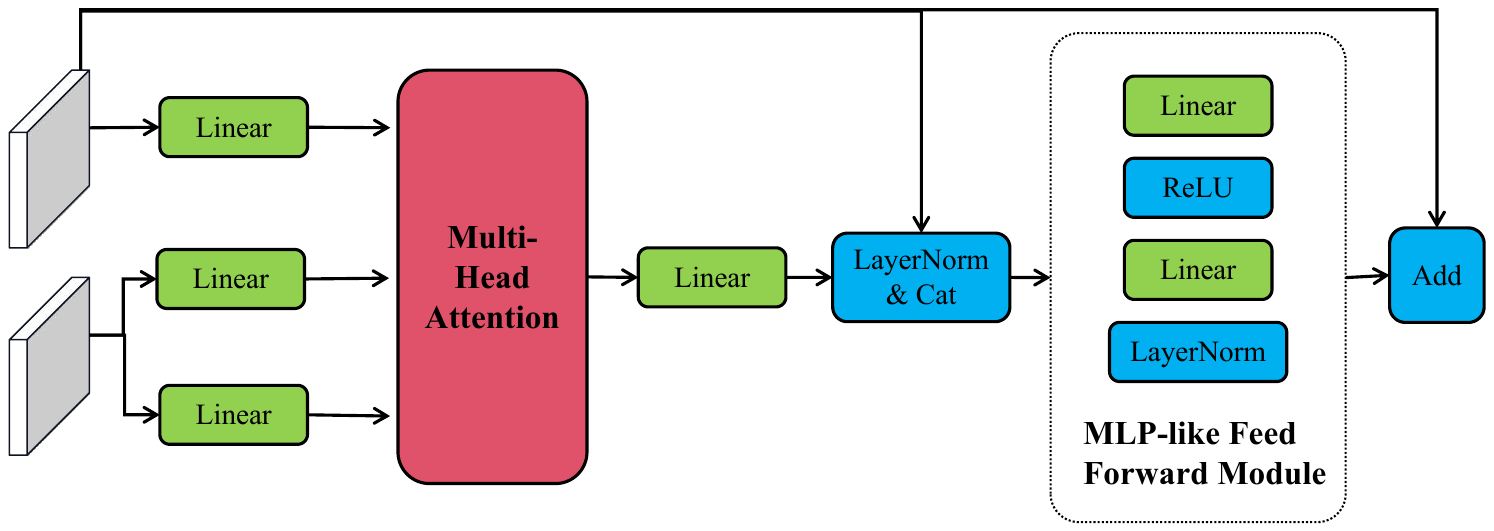}
  \caption{The architecture of self and cross attention of sub-pixel refinement module.}
  \label{fig:LoFTR}
\end{figure}

%% file: figure/feature_matching.tex
\begin{figure*}[htbp]
  \centering
  \includegraphics[width=0.8\textwidth]{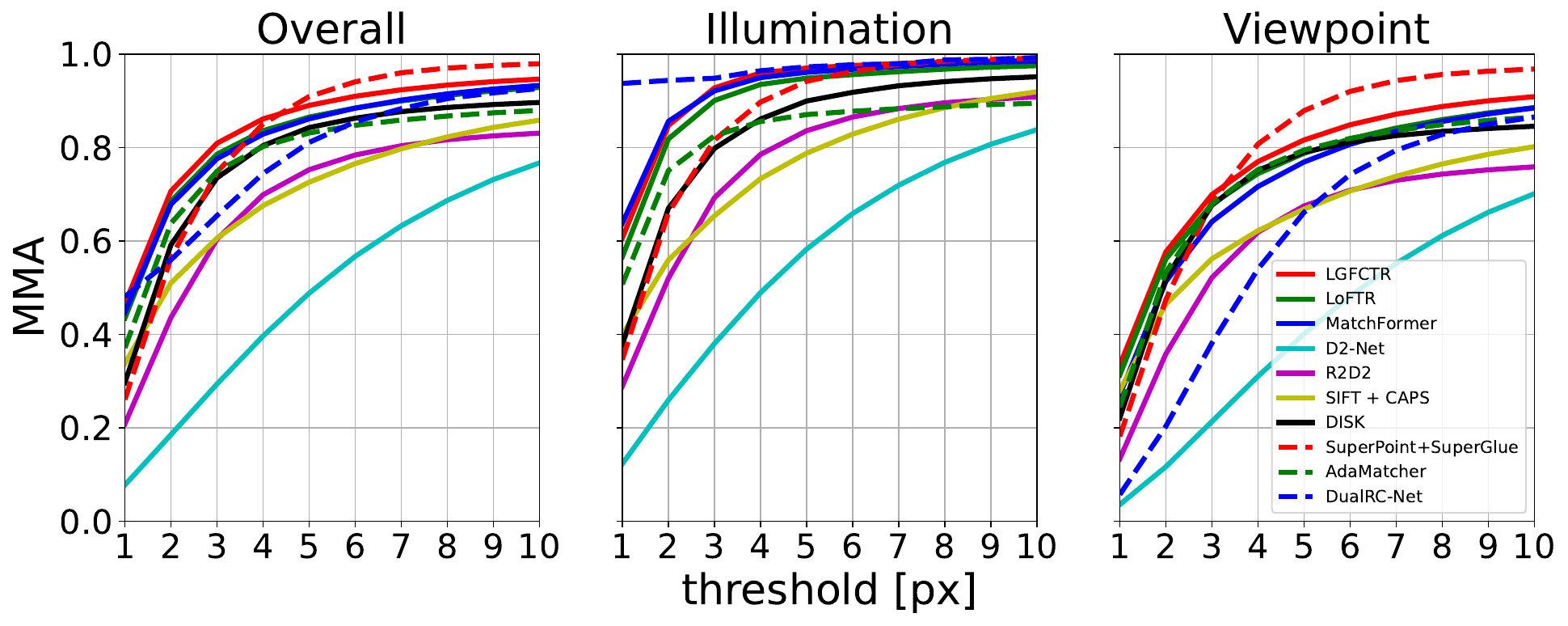}
  \caption{Mean matching accuracy results of feature matching on HPatches dataset.}
  \label{fig:feature_matching}
\end{figure*}

%% file: table/feature_matching.tex
\begin{table}[htbp]
\centering
\resizebox{\linewidth}{!}
{
\begin{tabular}{@{}ccccc@{}}
\toprule
\multirow{2}{*}{Category}       & \multirow{2}{*}{Method} & Illumination    & Viewpoint       & Overall         \\ \cmidrule(l){3-5} 
                                &                         & \multicolumn{3}{c}{MMA scores}                      \\ \midrule
\multirow{5}{*}{Detector-based} & D2-Net\cite{D2-Net}              & 0.5181          & 0.3652          & 0.4388          \\
                                & R2D2\cite{R2D2}                & 0.7252          & 0.5662          & 0.6427          \\
                                & DISK\cite{DISK}                & 0.8023          & 0.6812          & 0.7395          \\
                                & SIFT\cite{sift} + CAPS\cite{CAPS}      & 0.7237          & 0.6110          & 0.6652          \\
                                & SP\cite{SP} + SG\cite{SG}         & 0.8252          & \textbf{0.7356} & 0.7787    \\ \midrule
\multirow{5}{*}{Detector-free}  & DualRC-Net\cite{DualRC-Net}          & \textbf{0.9658} & 0.5412          & 0.7456          \\
                                & AdaMatcher\cite{AdaMatcher}          & 0.8079          & 0.6925          & 0.7481          \\
                                & MatchFormer\cite{MatchFormer}         & {\ul0.9065}          & 0.6840          & 0.7911          \\
                                & LoFTR\cite{LoFTR}               & 0.8827          & 0.7072          & {\ul0.7917}          \\
                                & LGFCTR                  & 0.9062          & {\ul 0.7315}    & \textbf{0.8156} \\ \bottomrule
\end{tabular}
}
\caption{MMA scores of feature matching on HPatches dataset.}
\label{tab:feature_matching}
\end{table}

%% file: figure/HPatches_i.tex
\begin{figure*}[htbp]
  \centering
  \includegraphics[width=0.95\textwidth]{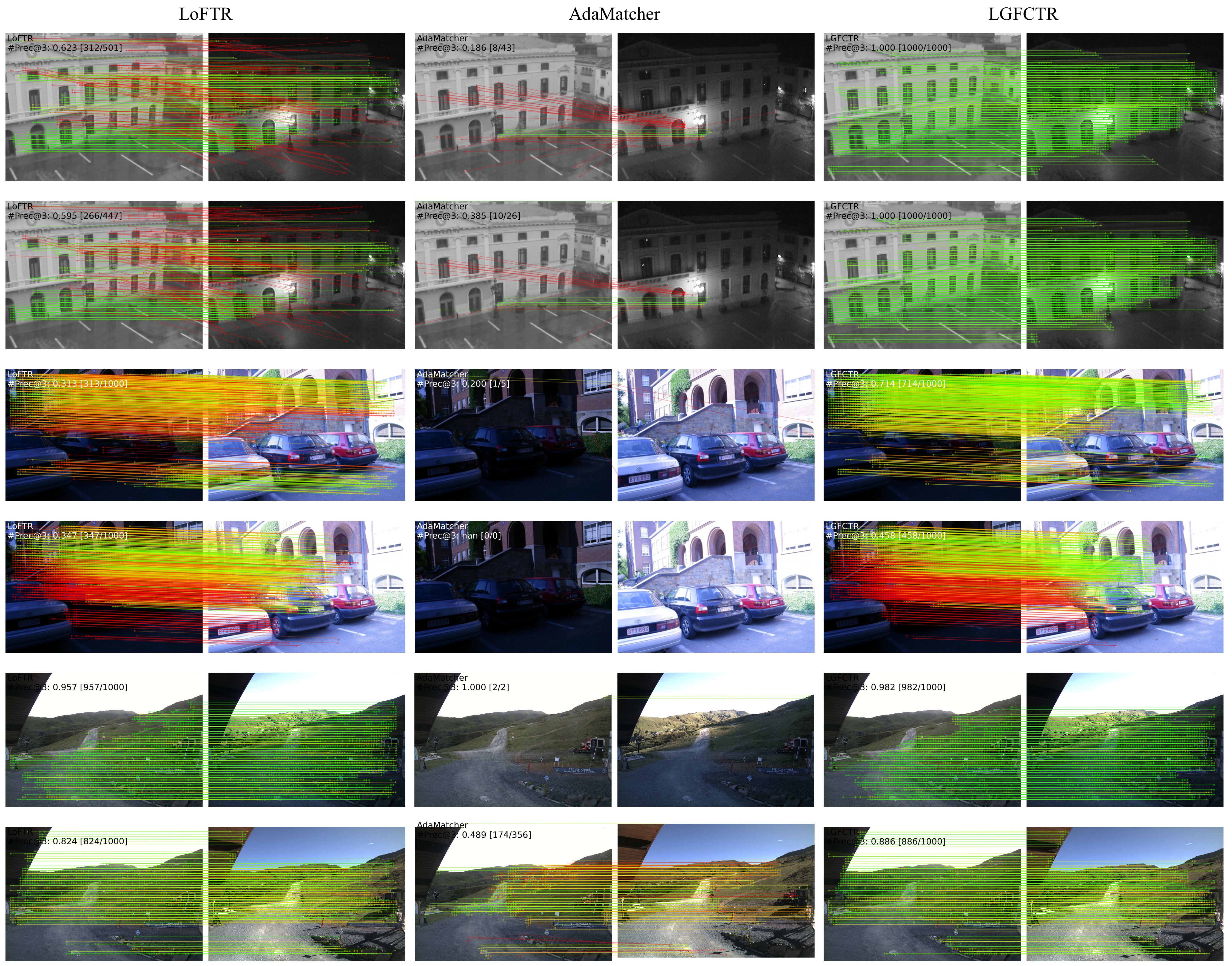}
  \caption{The qualitative results on several illumination pairs of HPatches dataset. The top 1000 matches with highest confidences are drawn and colored according to the threshold of 3px. The green and red lines represent less and more matching errors, respectively. And the matching accuracy is annotated on the top left of each pair of images.}
  \label{fig:HPatches_i}
\end{figure*}

%% file: figure/HPatches_v.tex
\begin{figure*}[htbp]
  \centering
  \includegraphics[width=0.95\textwidth]{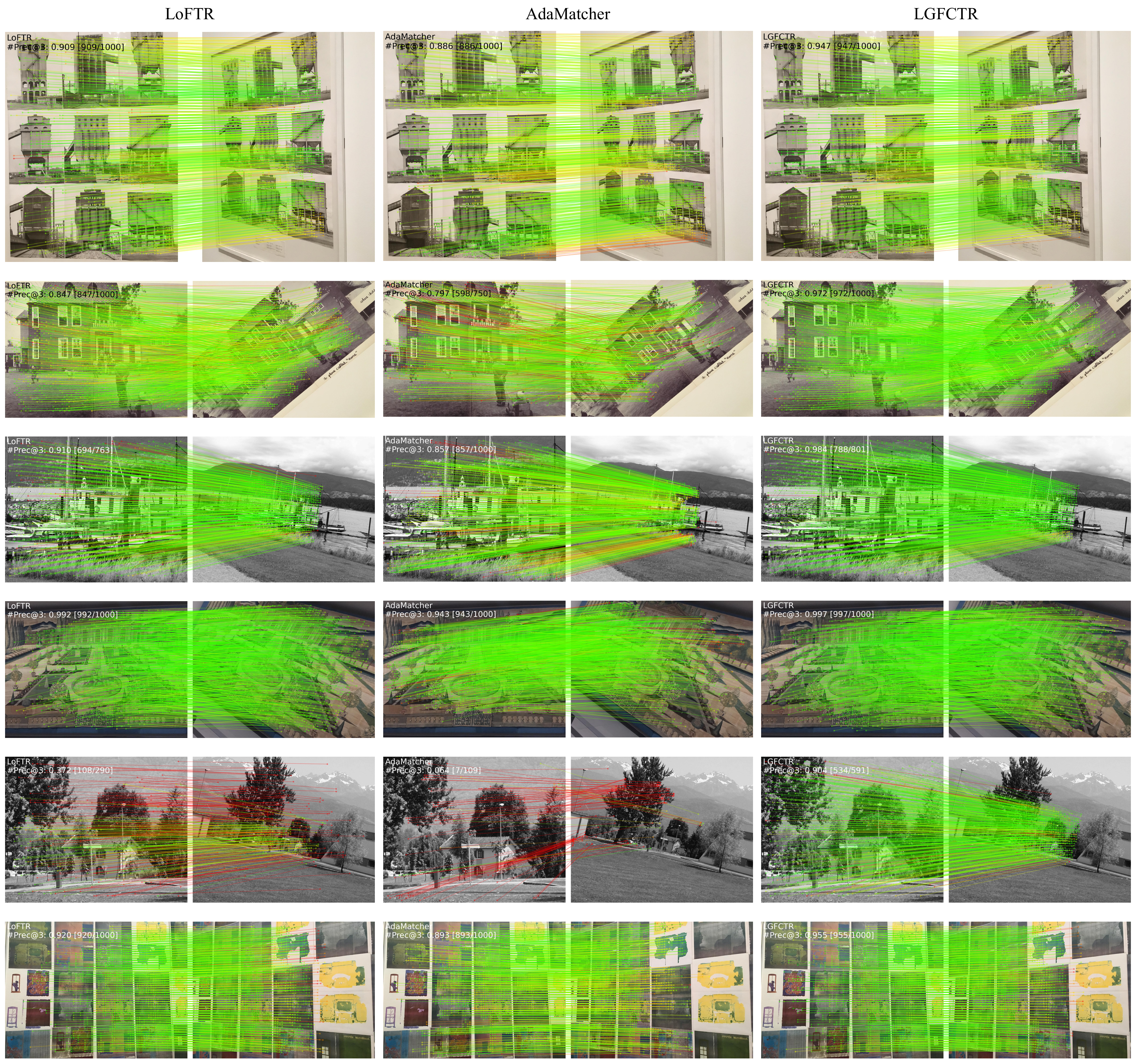}
  \caption{The qualitative results on several viewpoint pairs of HPatches dataset.}
  \label{fig:HPatches_v}
\end{figure*}

%% file: figure/MegaDepth_suppl.tex
\begin{figure*}[htbp]
  \centering
  \includegraphics[width=0.95\textwidth]{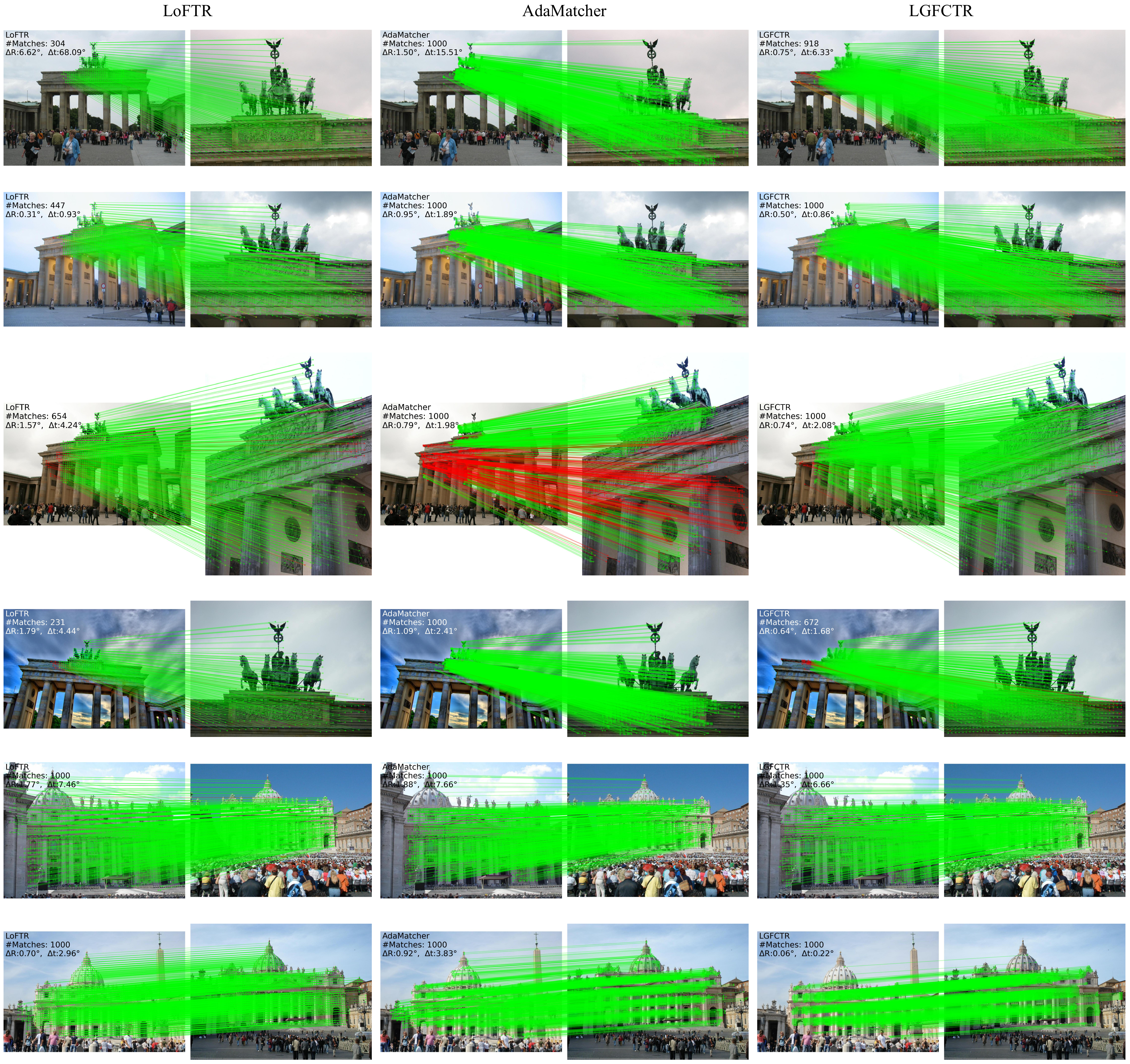}
  \caption{More qualitative results on MegaDepth dataset.}
  \label{fig:MegaDepth_suppl}
\end{figure*}

%% file: figure/subpixel_suppl.tex
\begin{figure*}[htbp]
  \centering
  \includegraphics[height=0.9\textheight]{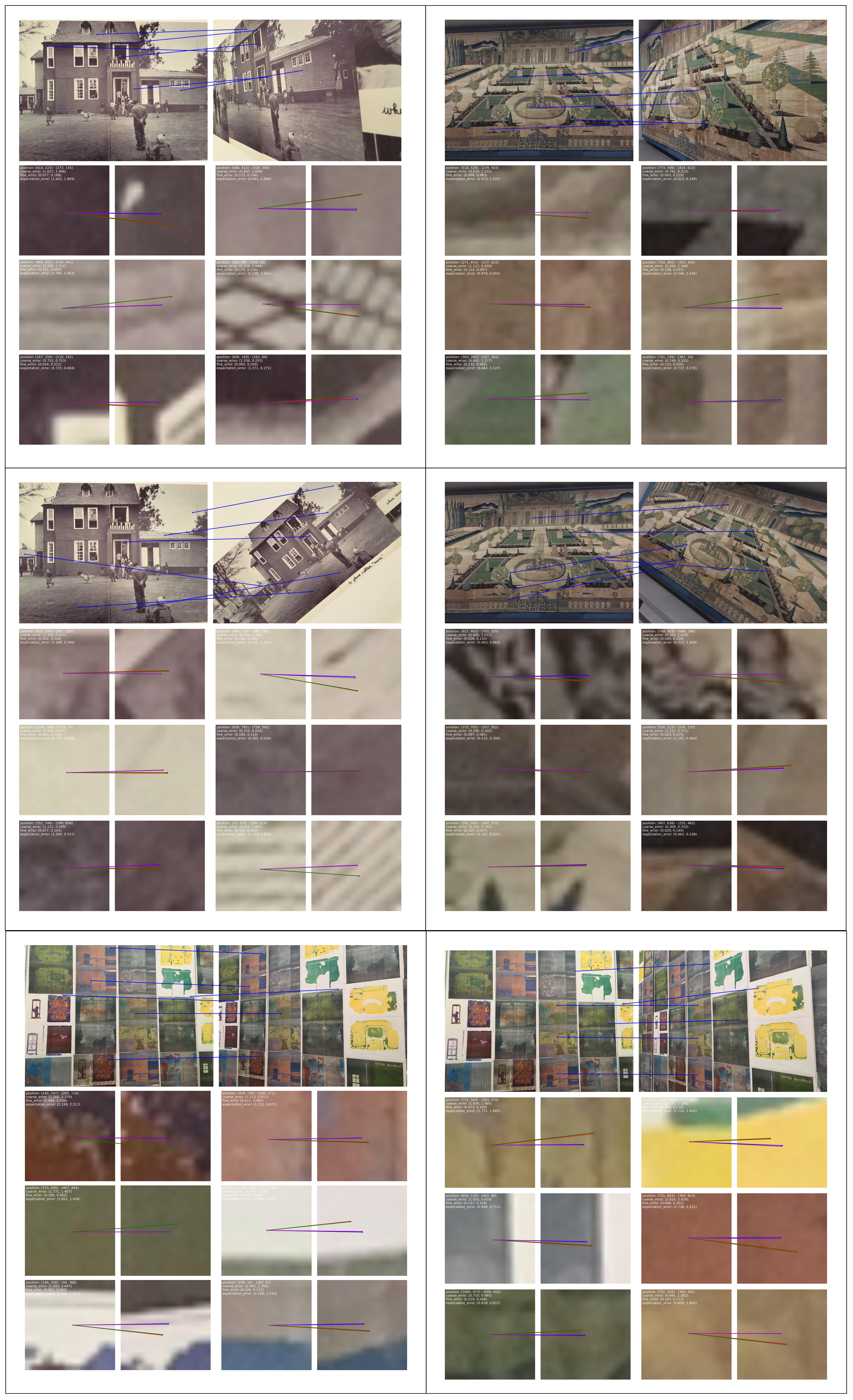}
  \caption{More visualizations of sub-pixel refinement module. For each pair of images, six pairs of zoomed-in patches are shown.}
  \label{fig:subpixel_suppl}
\end{figure*}

%% file: figure/MSA_suppl.tex
\begin{figure*}[htbp]
  \centering
  \includegraphics[width=0.95\textwidth]{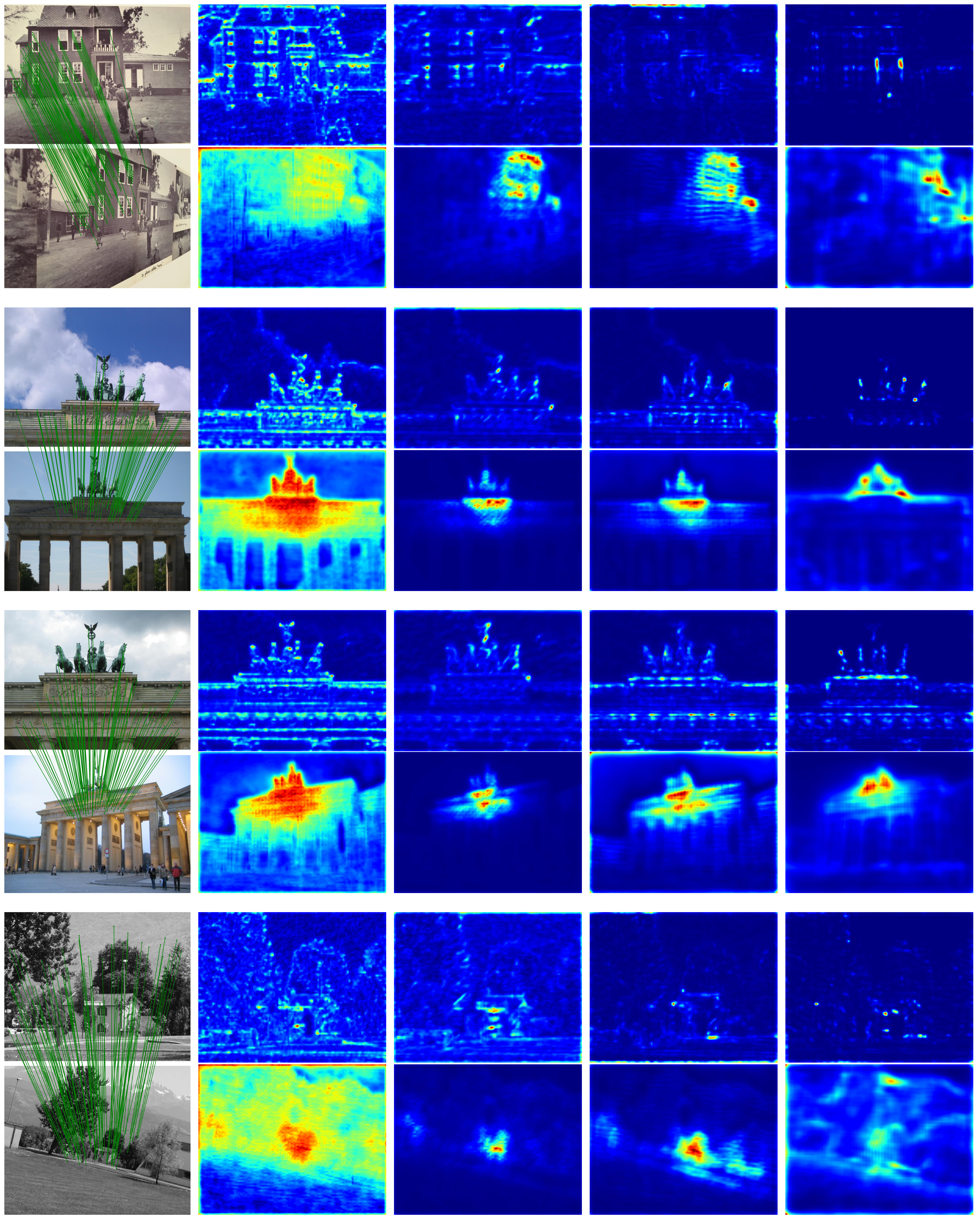}
  \caption{More visualizations of multi-scale attention.}
  \label{fig:MSA_suppl}
\end{figure*}